\documentclass[letterpaper]{article} 
\usepackage{aaai23}  
\usepackage{times}  
\usepackage{helvet}  
\usepackage{courier}  
\usepackage[hyphens]{url}  
\usepackage{graphicx} 
\usepackage{natbib}  
\usepackage{caption} 
\usepackage{algorithm}

\usepackage{bbm}
\usepackage{amsmath}
\usepackage{textcomp}
\usepackage{tabularx}
\usepackage{booktabs}
\usepackage{enumerate}
\usepackage{appendix}
\usepackage{multirow, makecell}

\usepackage[dvipsnames]{xcolor}
\usepackage{algpseudocode}
\definecolor{yellow}{RGB}{255, 204, 0}
\definecolor{red}{RGB}{234, 103, 12}

\usepackage{newfloat}
\usepackage{listings}
\DeclareCaptionStyle{ruled}{labelfont=normalfont,labelsep=colon,strut=off} 
\floatstyle{ruled}
\newfloat{listing}{tb}{lst}{}
\floatname{listing}{Listing}
\title{Two Wrongs Don't Make a Right: \\ Combating Confirmation Bias in Learning with Label Noise}
\author{
  Mingcai Chen, Hao Cheng, Yuntao Du, Ming Xu, Wenyu Jiang, Chongjun Wang\thanks{Corresponding authors}\\
}
\affiliations{
  State Key Laboratory for Novel Software Technology at Nanjing University\\
  Nanjing University, Nanjing 210023, China\\
  \{chenmc, chengh, duyuntao, lygjwy\}@smail.nju.edu.cn, xuming0830@gmail.com, chjwang@nju.edu.cn
}

\begin{document}
\maketitle
\begin{abstract}
    Noisy labels damage the performance of deep networks. 
    For robust learning, a prominent two-stage pipeline alternates between eliminating possible incorrect labels and semi-supervised training.
    However, discarding part of noisy labels could result in a loss of information, especially when the corruption has a dependency on data, e.g., class-dependent or instance-dependent.
    Moreover, from the training dynamics of a representative two-stage method DivideMix, we identify the domination of confirmation bias: pseudo-labels fail to correct a considerable amount of noisy labels, and consequently, the errors accumulate.
    To sufficiently exploit information from noisy labels and mitigate wrong corrections, we propose Robust Label Refurbishment (Robust~LR)---a new hybrid method that integrates pseudo-labeling and confidence estimation techniques to refurbish noisy labels.
    We show that our method successfully alleviates the damage of both label noise and confirmation bias.
    As a result, it achieves state-of-the-art performance across datasets and noise types, namely CIFAR under different levels of synthetic noise and Mini-WebVision and ANIMAL-10N with real-world noise. 
\end{abstract}

\section{Introduction}
Given certain capacity, deep networks have the capability of fitting arbitrary complex functions \cite{cybenko1989approximation}. 
However, the randomization tests on common architectures \cite{edgington2007randomization,zhang2016understanding,arpit2017closer} show that they also easily fit training data with random labels.
This phenomenon naturally raises the question of how deep learning continues to succeed in the presence of label noise.

\begin{figure*}[t] 
    \centering
    \includegraphics[width=\textwidth]{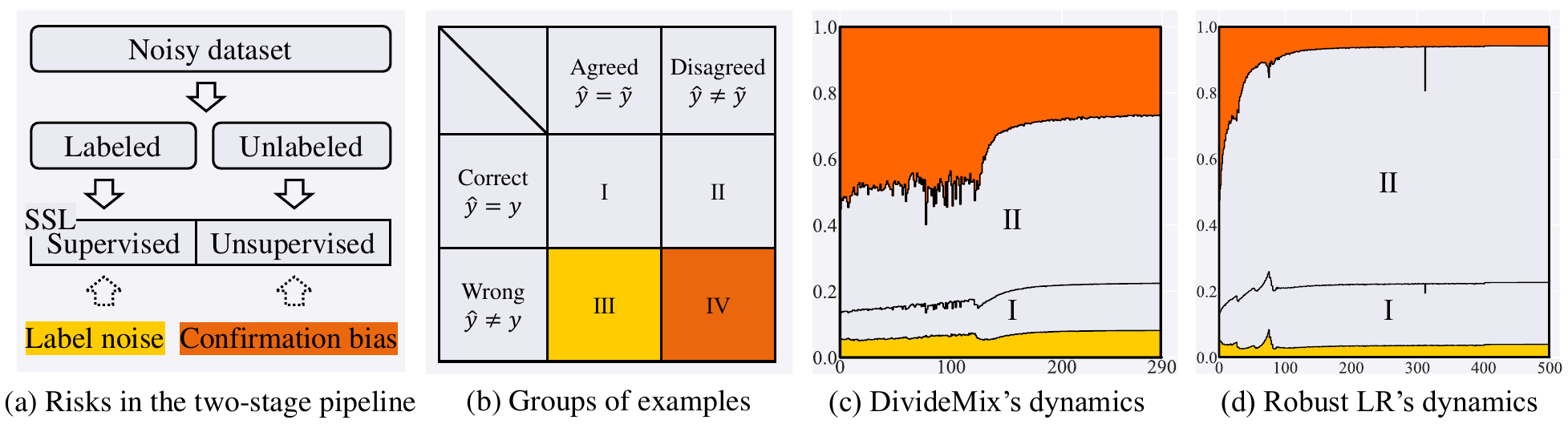} 
    \caption{
    Two-stage pipeline fails to correct a large proportion of wrong labels, evidenced by the training dynamics.
    Underlying ground-truth label, noisy label, and predicted label are denoted as $y$, $\tilde{y}$, $\hat{y}$ respectively.
    In every epoch, the examples are divided into four groups as shown in (b):
    I. The predicted labels agree with the clean labels. 
    II. The predicted labels correct the noisy labels. 
    III. The predicted labels agree with the noisy labels. 
    IV. The predicted labels fail to correct the given labels. 
    In (c) and (d), the x-axis denotes the epoch, and the y-axis denotes the proportion of different groups.
    Best viewed in color.
    }
    \label{motivation}
\end{figure*}

Recently, the state-of-the-art two-stage methods have significantly improved noise robustness by incorporating Semi-Supervised Learning (SSL) \cite{ding2018semi,nguyen2019self,li2020dividemix,zhou2021robust}.
The pipeline of a representative algorithm DivideMix \cite{li2020dividemix} is shown in Fig.~\ref{motivation}(a). 
In the first stage, problematic labels are identified and removed according to the per-example loss, i.e., the so-called  ``small-loss trick''.
Therefore, the noisy dataset is divided into a labeled subset and an unlabeled subset. 
In the second stage, DivideMix calls an SSL algorithm named MixMatch \cite{berthelot2019mixmatch}, which minimizes the entropy of predictions on unlabeled examples through pseudo-labels.
Such a pipeline leverages mislabeled data, improving the robustness to heavy and complex label noise.

However, we conclude that the two-stage pipeline suffers from two drawbacks.
On the one hand, according to Vapnik's principle \cite{Vapnik1998,Chapelle2006},\footnote{When solving a problem of interest, do not solve a more general problem as an intermediate step.} discarding possible noisy labels to construct an SSL setting is inefficient. 
Specifically, some correct labels are wrongly filtered. 
What's more, incorrect labels may also contain knowledge about the targets \cite{yu2018learning,ishida2017learning,kim2019nlnl,berthon2021confidence}.
For example, when an airplane image is mislabeled as a bird, the noisy label encodes the similarity information between the object of interest and the ``bird'' class.
On the other hand, when introducing pseudo-labels during the SSL stage, confirmation bias \cite{tarvainen2017mean,arazo2020pseudo} appears:
Those confident but wrong predictions would be used to guide subsequent training, leading to a loop of self-reinforcing errors.
\emph{Label noise, together with confirmation bias, damage the performance.}

To observe the erroneous pseudo-labeling, we draw the training dynamics of a recent two-stage method DivideMix \cite{li2020dividemix} on the corrupted training set of CIFAR-10 \cite{krizhevsky2009learning} (under 90\% synthetic symmetric noise).
In every epoch, examples are grouped according to the relationship between their predicted labels, corrupted labels, and underlying ground-truth labels as in Fig.~\ref{motivation}(b).
The \textcolor{yellow}{yellow color} indicates the examples whose predicted labels agree with given noisy labels, i.e., III. predicted label $=$ noisy label $\neq$ ground-truth.
The small \textcolor{yellow}{yellow region} at the bottom of Fig.~\ref{motivation}(c) suggests that the model only agrees with a small fraction of noisy labels.
It's because DivideMix would filter possible wrong labels and avoid fitting them.
On the other side, the \textcolor{red}{red color} indicates those predictions which fail to correct the noisy labels, i.e., IV. predicted label $\neq$ noisy label and predicted label $\neq$ ground-truth.
From the \textcolor{red}{red region} at the top of Fig.~\ref{motivation}(c), incorrect corrections comprise a large part throughout the training process.
Considering the wrong pseudo-labels would be used for self-training, it causes the confirmation bias problem, affecting performance adversely.

Our work begins by suggesting that better robustness can be achieved by sufficiently exploiting the information in the noisy labels and mitigating the side-effect of SSL.
We observe one of the recent two-stage methods as Fig.~\ref{motivation}(c):
The pseudo-labels dominate the given noisy labels during training.
We propose a hybrid method named Robust LR to address the problem. 
It estimates the label confidence by modeling the per-example loss and then accordingly refurbishes noisy labels through a dynamic convex combination with pseudo-labels. 
Robust LR improves upon the two-stage pipeline by leveraging all noisy labels and constructing target labels in a more fine-grained manner.
To further alleviate confirmation bias: 1). Two models are trained simultaneously, where each model interacts with its peer through pseudo-labeling and confidence estimation. 2). Different augmentation strategies are deployed for loss modeling and learning following recent findings \cite{chen2020simple,nishi2021augmentation}.
For comparison, we draw Fig.~\ref{motivation}(d) using our method under the same setting. 
Compared with Fig.~\ref{motivation}(c), the \textcolor{red}{red region}, which indicates wrong corrections, are much smaller.
It shows that our approach alleviates the damage of wrong pseudo-labels while combating label noise. 
To sum up, we highlight the contributions of this paper as follows:
\begin{itemize}
    \item We analyze the inefficiency of the two-stage pipeline and suggest that there is a loss of information when transforming the label noise problem into SSL.
    Moreover, the visualization of the training dynamics helps us identify the domination of confirmation bias (see Fig.~\ref{motivation}(c)).
    \item To address this, we propose a hybrid method named Robust~LR.
    By integrating pseudo-labeling and confidence estimation techniques into label refurbishment, it successfully leverages all noisy labels and alleviates the damage of both label noise and confirmation bias.
    \item We experimentally show that our method advances state-of-the-art results on CIFAR with synthetic label noise, as well as the real-world noisy dataset Mini-WebVision and ANIMAL-10N.
    Besides, we systematically study the components of Robust LR to examine their impacts.
\end{itemize}

\section{Related work}

The label noise is ubiquitous in real-world data.
When the noise rate is insignificant, it can be implicitly dealt with.
For example, the noise labels in MNIST, CIFAR, and ImageNet (some of them are reported in \url{https://labelerrors.com/}), are usually neglected.
Regularization techniques, including Dropout \cite{srivastava2014dropout}, weight decay \cite{krogh1992simple}, and the inherent robustness in deep networks \cite{zhang2016understanding} combat label noise.

The damage of noisy labels gradually appears as noise becomes non-negligible.
Some methods assume a class-dependent (or instance-independent) label noise, i.e., the distribution of noisy labels only dependent on the ground-truth label:
\begin{equation}
    p(\tilde{y}=j\mid y=i,X=x)=p(\tilde{y}=j\mid y=i)
\end{equation}
The corruption process thus can be modeled by a label transition matrix $T\in [0,1]^{C \times C}$ where $T_{ij}:=p(\tilde{y}=j\mid y=i)$ and $C$ is the number of classes.
Webly learning \cite{chen2015webly} adds an extra noise adaptation layer on top of the base model to mimic the transition behavior.
The base model is first trained on easy examples, and then the entire model is trained on the noisy dataset.  
Backward correction \cite{patrini2017making} estimates the label transition through the outputs of a network trained on the noisy dataset.
Then it trains another network with weighted loss, where the weights are from the estimated label transition matrix.
Forward correction \cite{patrini2017making} does the same to obtain the matrix. 
But it instead corrects the outputs during forward pass when trains a new network.
To better estimate the transition matrix, Dual T \cite{yao2020dual} factorizes it into two easy-to-estimate matrices.
The effectiveness of these approaches depends on whether the transition matrix is accurate. 
Besides, the noise type could be more complex in real-world, e.g., instance-dependent:
\begin{equation}
    p(\tilde{y}=j\mid y=i,X=x)=T_{i,j}(x)p(y=j\mid y=i)
\end{equation}
where $T_{i,j}(x)$ is the instance-dependent noise model.
The aforementioned methods have difficulty in modeling such complex noise.

A large part of the methods achieves robustness by relying on the internal noise tolerance of deep networks.
They mainly differ in the example selection, loss weighting, or label refurbishment strategies \cite{frenay2013classification,algan2021image,DBLP:conf/icml/SongK019}.
Bootstrapping \cite{reed2014training} uses the interpolation of labels and model predictions for training.
Decouple \cite{malach2017decoupling} updates two predictors with only disagreed examples.
Activate bias \cite{chang2017active} emphasizes high variance examples.
MentorNet \cite{jiang2018mentornet} weights examples using a pre-trained teacher network.
Co-teaching \cite{han2018co} maintains two models where one selects examples with small losses to update another.
Based on Co-teaching, Co-teaching+ \cite{yu2019does} prevents two models from converging to a consensus by only considering disagreed examples.
D2L \cite{ma2018dimensionality} adopts a measure called local intrinsic dimensionality.
Labels are refurbished to prevent the increase of intrinsic dimension.
SELFIE \cite{DBLP:conf/icml/SongK019} only considers examples with consistent predictions for refurbishment.
TopoFilter \cite{wu2020topological} adopts a different selection criteria by exploring the latent representational space.
Self-adaptive training \cite{huang2020self} uses the exponential moving average of predictions as pseudo-labels.
SEAL \cite{chen2020beyond} retrains a model with the average predictions of a teacher model.
However, these methods may suffer from big performance drops under heavy noise due to inaccurate correction, weighting, or refurbishment. 

\begin{table}
    \includegraphics[width=0.95\columnwidth]{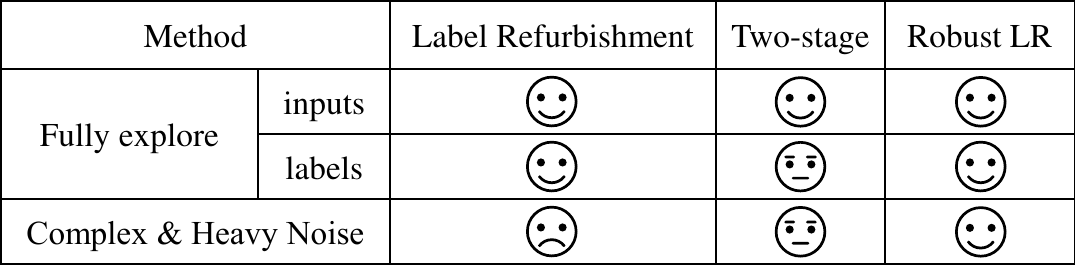} 
    \centering
    \caption{Comparison of training schemes.
    }
    \label{advantage}
\end{table}

Recently, the two-stage pipeline has gained much attention.
SELF \cite{nguyen2019self} first uses the ensemble of predictions to filter problematic labels.
In the second stage, it performs an SSL method named Mean Teacher \cite{tarvainen2017mean}.
DivideMix \cite{li2020dividemix} uses the Gaussian Mixture Model (GMM) to separate examples with small and big losses, and they are treated as clean and noisy examples, respectively.
Then the SSL method MixMatch \cite{berthelot2019mixmatch} is used to leverage the feature information.
RoCL \cite{zhou2021robust} selects clean examples according to the consistency of the loss and output, followed by a self-training method.
This type of method utilizes SSL to leverage mislabeled examples.
However, we suggest that they fail to exploit all noisy labels and suffer from wrong corrections.
The proposed method Robust LR leverages possible noisy labels.
It preserves label information in a soft manner by adopting successful ideas from the two-stage pipeline and SSL into the classic label refurbishment process, as shown in Table \ref{advantage}.
Furthermore, Robust LR is dedicated to alleviating confirmation bias. 
Different augmentation strategies and co-training are combined to form a hybrid method.

\section{Method}
\subsection{Overview of Robust LR}

\begin{figure}
\centering
\includegraphics[width=0.83\columnwidth]{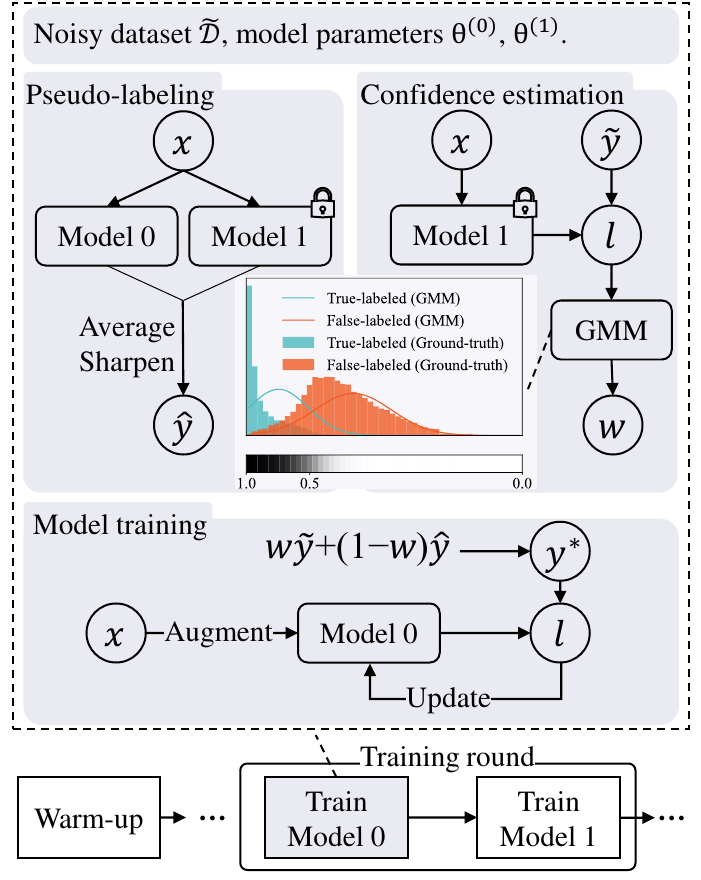}
    \caption{Pipeline of Robust LR.
    }
    \label{fig_Robust LR}
\end{figure}

Robust LR refurbishes the noisy labels before training.
To reduce the marginalized effect of wrong labels, the refurbished label $y^*\in  \Delta ^{C-1}$ (where $\Delta ^{C-1}$ is the probability simplex) comes from a dynamic convex combination of the noisy label $\tilde{y}$ (one-hot label over $C$ classes) and the soft pseudo-label $\hat{y}$ (predicted probability distribution over $C$ classes).
\begin{equation}  \label{refur_eq}
    y^*=w\tilde{y}+(1-w)\hat{y}
\end{equation}
The pseudo-label $\hat{y}$ is obtained from the models' prediction.
The weight $w$, i.e., the clean probability, is estimated using a two-component GMM fitted on the per-example loss.
To further alleviate confirmation bias, two models are simultaneously trained, where one model contributes to another's confidence estimation and pseudo-labeling process.
They have the same structure but different parameters $\theta^{(0)},\theta^{(1)}$.
The overall pipeline of Robust LR is shown in Fig.~\ref{fig_Robust LR} and Algorithm \ref{alg}. 
In every training round, the confidence estimation and pseudo-labeling are performed first. 
Then the model is trained with the refurbished labels. 

\subsection{Warm-up}
As shown in \cite{arpit2017closer}, deep models tend to fit clean examples first.
Therefore, Robust LR warms two models up by shortly training them on the noisy dataset. 
The commonly used mini-batch gradient descent algorithm is performed to update the parameters.
For illustration, we denote this process as Train(dataset, parameters, number of iterations). 
Thus, the warm-up process is:
\begin{equation} 
    \text{Train}(\tilde{\mathcal{D}},\theta^{(m)},I_{warm})\quad \text{for } m=0,1
\end{equation}
where $I_{warm}$ is a small number of iterations so that the training ends before models fitting too many noisy labels. 

\subsection{Main training round}
\subsubsection{Confidence estimation}
It has been shown that models are prone to present smaller losses on clean examples \cite{arpit2017closer,chen2019understanding,han2018co,li2020dividemix}.
Therefore, Robust LR estimates the label confidence based on the loss value.
Specifically, the per-example cross-entropy loss $\text{H}$ between the noisy label and the prediction is first calculated,
\begin{equation} 
        \ell_i=\text{H}(\tilde{y}_i, \text{p}(y\mid x_i;\theta^{(1-m)}))
\end{equation}
Then a two-component one-dimensional GMM is used to model the distribution of per-example loss,
\begin{equation}
    \begin{aligned}
        \mathcal{W}&=\text{GMM}(\{(\ell_i)\}_{i=1}^{N})\\
    \end{aligned}
\end{equation}
where $\mathcal{W}=\{w_i\}^N_{i=1}$ is the label confidence which equals to the probability of each loss value belonging to the GMM component with a smaller mean.
The parameters of GMM are determined using the expectation-maximization algorithm. 
The procedure follows the standard practice,  so we don't elaborate on the details here.
Note that, to alleviate confirmation bias, the label confidence for the current model $m$ comes from the predictions of another model $1-m$. 

\subsubsection{Pseudo-labeling}
To correct the noisy labels with accurate pseudo-labels, two models' predictions are averaged and then sharpened,
\begin{equation}
    \begin{aligned}
        \hat{y}_i=\text{Sharpen}(\frac{\text{p}(y\mid x_i;\theta^{(m)})+\text{p}(y\mid x_i;\theta^{(1-m)})}{2})\\
    \end{aligned}
\end{equation}
where the sharpening function scales the categorical distribution with a hyper-parameter $T$,
\begin{equation}
    \text{Sharpen}(p)_i=\frac{p_i^{\frac{1}{T}}}{\sum_{j=1}^{C}p^{\frac{1}{T}}_{j}}
\end{equation}
where $C$ is the number of classes. $p_i$ is the probability of $i$-th class of input distribution $p$.

\begin{algorithm}
    \caption{Robust LR} 
    \label{alg}
    \textbf{Input}:
    Noisy dataset $\tilde{\mathcal{D}}=\{(x_i,\tilde{y}_i)\}^{N}_{i=1}$,
    \# iterations for warm up $I_{warm}$, 
    \# iterations in main training round $I$, 
    \# training rounds $R$, 
    training strategy $\text{Train}(\text{dateset},\text{parameters},\text{\# iterations})$.
    \\
    \textbf{Output}: model's parameters $\theta^{(0)}$, $\theta^{(1)}$ 
    \begin{algorithmic}[1] 
    \State Randomly initialize $\theta^{(0)}$, $\theta^{(1)}$
    \State $\text{Train}(\tilde{\mathcal{D}},\theta^{(m)},I_{warm}) \quad \text{for } m=0, 1$
        \Comment{warm up}
    
    \For{$r=1$ to $R$} 
        \For{$m=0$ to $1$}
            \Comment{train two models separately}
            \For{$i=0$ to $N$}
                \State $\ell_i=\text{H}(y_i, \text{p}(y\mid x_i;\theta^{(1-m)}))$
                \State \Comment{obtain per-example loss}
            \EndFor
            \State $W=\text{GMM}(\{(\ell_i)\}_{i=1}^{N})$
                \Comment{fit GMM}
            \For{$i=0$ to $N$}
                \State $\hat{y}_i=\text{Sharpen}(\frac{\text{p}(y\mid x_i;\theta^{(m)})+\text{p}(y\mid x_i;\theta^{(1-m)})}{2})$
                \State \Comment{pseudo-label}
                \State $y^*_i=w_i\tilde{y}_i+(1-w_i)\hat{y}_i$ \Comment{refurbish}
            \EndFor
            \State $\text{Train}(\{(\text{Aug}(x_i),y_i^*)\}_{i=1}^N,\theta^{(m)},I)$
        \EndFor
    \EndFor

    \end{algorithmic} 
\end{algorithm}

\subsubsection{Model training}
After label refurbishment using the estimated confidence and pseudo-labels according to Equation~\ref{refur_eq}, current model $m$ is trained with the refurbished labels for $I$ iterations,
\begin{equation}
    \begin{gathered}
        \text{Train}(\{(\text{Aug}(x_i),y_i^*)\}_{i=1}^N,\theta^{(m)},I)
    \end{gathered}
\end{equation}
where $\text{Aug}(\cdot)$ is the data augmentation function introduced in the next section.
The cross-entropy between the soft labels and predictions is used as loss fucntion here.
After the training of model $m$, another model $1-m$ is trained similarly.
This process proceeds until reaching a fixed number of training rounds.

During implementation, a regularization loss term is used as in  \cite{tanaka2018joint,arazo2019unsupervised,li2020dividemix}. 
It encourages the network to output uniform distribution across examples in the mini-batch.
\begin{equation}
\begin{aligned}
    L_{reg}&=\sum_c \pi_c log(\frac{\pi_c}{\Bar{p}_c})\\
    \Bar{p}_c&=\frac{1}{B}\sum_{i=1}^B \text{p}(y=c\mid x_i;\theta)
\end{aligned}
\end{equation}
where $\pi$ is the uniform prior distribution, we set $\pi_c=\frac{1}{C}$. 

For asymmetric noise, we add a negative entropy loss term during warm-up following \cite{pereyra2017regularizing,li2020dividemix}.
\begin{equation}
\sum_c p(y\mid x;\theta)log(p(y\mid x;\theta))
\end{equation}

\subsection{The different augmentation strategies} \label{sectionaug}

\begin{table*}
    \centering
    \begin{tabular}{lr|c|c|c|c|c||c|c|c|c}
        \toprule
		Dataset               &      &\multicolumn{5}{c||}{CIFAR-10}& \multicolumn{4}{c}{CIFAR-100}\\\midrule
		Noise type &      &\multicolumn{4}{c|}{Sym.}& \multicolumn{1}{c||}{Asym.} & \multicolumn{4}{c}{Sym.}\\\midrule
		\multicolumn{2}{l|}{Method/Noise ratio}        & 20\% & 50\% & 80\% & 90\% & 40\% & 20\% & 50\% & 80\% &  90\% \\ \midrule
		\multirow{2}{*}{F-correction \cite{patrini2017making}}          & Best  & 86.8 & 79.8 & 63.3 & 42.9 & 87.2 & 61.5 & 46.6 & 19.9 & 10.2 \\
		                                       & Last  & 83.1 & 59.4 & 26.2 & 18.8 & 83.1 & 61.4 & 37.3 &  9.0 &  3.4 \\ \midrule
		\multirow{2}{*}{Co-teaching+ \cite{yu2019does}}          & Best  & 89.5 & 85.7 & 67.4 & 47.9 &  -   & 65.6 & 51.8 & 27.9 & 13.7 \\
		                                       & Last  & 88.2 & 84.1 & 45.5 & 30.1 &  -   & 64.1 & 45.3 & 15.5 &  8.8 \\ \midrule
		\multirow{2}{*}{P-correction \cite{yi2019probabilistic}}          & Best  & 92.4 & 89.1 & 77.5 & 58.9 & 88.5 & 69.4 & 57.5 & 31.1 & 15.3 \\
		                                       & Last  & 92.0 & 88.7 & 76.5 & 58.2 & 88.1 & 68.1 & 56.4 & 20.7 &  8.8 \\\midrule
		\multirow{2}{*}{Meta-Learning \cite{li2019learning}}         & Best  & 92.9 & 89.3 & 77.4 & 58.7 & 89.2 & 68.5 & 59.2 & 42.4 & 19.5 \\
		                                       & Last  & 92.0 & 88.8 & 76.1 & 58.3 & 88.6 & 67.7 & 58.0 & 40.1 & 14.3 \\\midrule
		\multirow{2}{*}{M-correction \cite{arazo2019unsupervised}}          & Best  & 94.0 & 92.0 & 86.8 & 69.1 & 87.4 & 73.9 & 66.1 & 48.2 & 24.3 \\
		                                       & Last  & 93.8 & 91.9 & 86.6 & 68.7 & 86.3 & 73.4 & 65.4 & 47.6 & 20.5 \\\midrule
		\multirow{2}{*}{DivideMix  \cite{li2020dividemix}}             & Best  & 96.1 & 94.6 & 93.2 & 76.0 & 93.4 & 77.3 & 74.6 & 60.2 & 31.5 \\
				                               & Last  & 95.7 & 94.4 & 92.9 & 75.4 & 92.1 & 76.9 & 74.2 & 59.6 & 31.0 \\\midrule
		\multirow{2}{*}{AugDesc$^*$ \cite{nishi2021augmentation}}        & Best  & 96.1 &  -   &   -  & 89.6 &  -   & 78.1 &  -   &  -   & 36.8 \\
				                               & Last  & 96.0 &  -   &   -  & 89.4 &  -   & 77.8 &  -   &  -   & 36.7 \\\midrule
		\multirow{2}{*}{Ours}            & Best & \textbf{96.5} & \textbf{95.8} & \textbf{94.3} & \textbf{92.8} & \textbf{94.4} & \textbf{79.1} & \textbf{75.3} & \textbf{66.7} &  \textbf{37.5} \\
				              & Last & \textbf{96.4} & \textbf{95.7} & \textbf{94.2} & \textbf{92.8} & \textbf{93.7} & \textbf{78.6} & \textbf{74.6} & \textbf{66.2} & \textbf{37.3} \\
        \bottomrule
    \end{tabular}
    \caption{Comparison with state-of-the-art methods on CIFAR10 and CIFAR-100 with synthetic noise. Sym. and Asym. are symmetric and asymmetric for short, respectively. 
    The results of other methods are from \cite{li2020dividemix}.
    The best results are indicated in bold. *AugDesc uses the same augmentation technique (RandAugment) as our method.}
    \label{table_CIFAR}
\end{table*}

Due to the lack of accurate supervised information, improving the generalization ability is the core task of learning with label noise.
Data augmentation is a common technique that approaches such a problem via applying stochastic transformation on images.

In Robust LR, forward pass serves three purposes: loss modeling, pseudo-labeling, and learning.
We use basic image augmentation for loss modeling and pseudo-labeling but stronger augmentations for learning.
This design is based on two recent findings:
1). In learning with label noise, using different augmentations for loss modeling and learning is more effective \cite{nishi2021augmentation}.
2). Unsupervised learning benefits from stronger data augmentation \cite{chen2020simple}, and we find the same preference can also be extended to this problem.

In particular, the basic image augmentation for loss modeling and pseudo-labeling consists of random crop and random horizontal flip.
The strong transformation $\text{Aug}(\cdot)$ consists of RandAugment \cite{cubuk2020randaugment} and Cutout \cite{devries2017improved}. 
RandAugment first randomly selects a given number of operations from a pre-defined set of transformations.
The set consists of geometric and photometric transformations, such as affine transformation and color adjustment. 
In the next, these operations are applied with given magnitudes.
Cutout randomly masks out square regions of images.
These augmentations are sequentially applied to the input images.
The settings of RandAugment are reported in the supplementary material.

\section{Experiment}
\subsection{Comparison with state-of-the-art methods}
We benchmark the proposed method on experimental settings using CIFAR-10, CIFAR-100 \cite{krizhevsky2009learning} with different levels of synthetic noises, as well as the real-world noisy dataset Mini-WebVision \cite{li2017webvision},  ANIMAl-10N \cite{DBLP:conf/icml/SongK019}.
\subsubsection{Synthetic label noise on CIFAR-10, CIFAR-100}

Following previous methods \cite{kim2019nlnl,li2020dividemix}, two types of synthetic noises are experimented: symmetric and asymmetric noise.
Symmetric noise is generated by assigning examples to random classes with the same probability.
The noise rate ranges from 20\% to 90\% (note that the noise labels are randomly distributed throughout $C$ classes, and the true labels may be maintained after corruption).
Asymmetric noise is generated by randomly corrupting labels according to a pre-defined transition matrix. 
Examples would only be corrupted to similar classes, such as deer to horse. 
40\% asymmetric noise is experimented (50\% being indistinguishable).

We report the average performance of Robust LR over 3 trials with different random seeds for generating noise and parameters initialization.
The backbone structure is PreAct Resnet \cite{he2016deep}. 
The training details are reported in the supplementary material.
Following previous work, the best test accuracy across all epochs and the averaged test accuracy over the last 10 epochs are both reported.
A validation set with 5,000 examples is drawn from the noisy training set for hyper-parameters tuning.
We find that two main hyper-parameters in Robust LR, namely temperature value and the weight for regularization term \cite{tanaka2018joint,arazo2019unsupervised}, don't need to be heavily tuned. 
Specifically, there are only two sets of hyper-parameters for light and heavy noise, respectively.
For light noise, namely CIFAR-10 under 20\% to 80\% symmetric noise, 40\% asymmetric noise, and CIFAR-100 under 20\% symmetric noise, the temperature is 1, and the coefficient for the regularization term is 2.
For heavy noise, namely CIFAR-10 under 90\% symmetric noise and CIFAR-100 under 50\% to 90\% symmetric noise, the temperature is $1/3$, and the coefficient for regularization term is 10.

As shown in Table~\ref{table_CIFAR}, our method consistently outperforms previous best results on all the settings. 
The improvement is substantial, especially when the noise is heavy.
For example, Robust LR obtains 92.8\% accuracy on CIFAR-10 under 90\% noise, surpassing the previous best by more than 3\%.
We remark that previous methods underperform under heavy noise because they fail to avoid confirmation bias.
It's worth noting that Robust LR outperforms AugDesc even with the same augmentation. 
It shows that our improvement also comes from other components.

\begin{figure}[h!]
    \centering
    \includegraphics[width=\columnwidth]{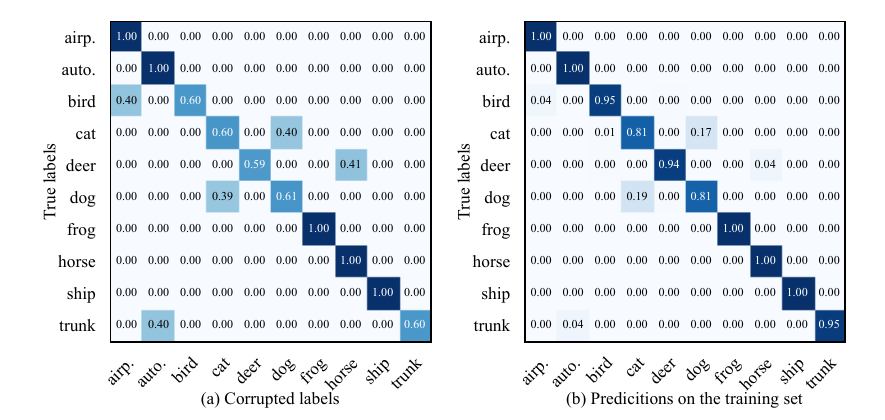} 
    \caption{Confusion matrices on CIFAR-10 under asymmetry noise. The airp. and auto. are airplane and automobile for short.} 
    \label{fig_confison}
\end{figure}

The distribution of asymmetric noise in the corrupted training set is shown in Fig.~\ref{fig_confison}.
The comparison between Robust LR and other methods is shown in Table~\ref{table_CIFAR}.
Our method outperforms the previous best method by over 1\%.
As we can see in Fig.~\ref{fig_confison}, Robust LR resists the mimicked class-dependent noise and correctly predicts most of them.

\subsubsection{Real-world label noise on Mini-WebVision and ANIMAl-10N}

\begin{table}
\centering
\begin{tabular}{l|cc|cc}
    \toprule
    \multirow{2}{*}{Method} & \multicolumn{2}{c|}{Mini-WebVision} & \multicolumn{2}{c}{ILSVRC12} \\ \cmidrule(r){2-3}\cmidrule(l){4-5}
                                & top-1  & top-5  & top-1  & top-5  \\ \midrule
        F-correction            & 61.12 & 82.68 & 57.36 & 82.36 \\
        Decoupling              & 62.54 & 84.74 & 58.26 & 82.26 \\
        D2L                     & 62.68 & 84.00 & 57.80 & 81.36 \\
        MentorNet               & 63.00 & 81.40 & 57.80 & 79.92 \\
        Co-teaching             & 63.58 & 85.20 & 61.48 & 84.70 \\
        Iterative-CV            & 65.24 & 85.34 & 61.60 & 84.98 \\
        DivideMix               & 77.32 & 91.64 & 75.20 & 90.84 \\ 
        Robust LR                      & \textbf{81.84} & \textbf{94.12} & \textbf{75.48} & \textbf{93.76} \\ \bottomrule
    \end{tabular}
\caption{Comparison with other methods on Mini-WebVision. The results of other methods are from \cite{li2020dividemix}.}
\label{table_webvision}
\end{table}

\begin{table}
\centering
\begin{tabular}{ccccc}
\toprule
SELFIE & PLC & NCT &  Robust LR\\ \midrule
81.8   & 83.4 & 84.1  &  \textbf{88.5} \\  \bottomrule
\end{tabular}
\caption{Comparison with other methods on ANIMAl-10N. The results of other methods are from \cite{DBLP:conf/cvpr/ChenSHS21}.}
\label{table_ANIMAl}
\end{table}

To verify the effectiveness of our approach on the real-world large-scale noisy dataset, we then conduct experiments on Mini-WebVision and ANIMAl-10N.
WebVision is crawled from Flickr and Google using the same 1,000 classes as the ImageNet ILSVRC12 dataset for querying.
The estimated noise rate is 20\%.
Following the setting of previous work \cite{chen2019understanding,li2020dividemix}:
The first 50 classes of the ImageNet ILSVRC12 dataset are compared, and its validation set is used.
In terms of the hyper-parameters, the temperature is 3, and the coefficient for the regularization term is 1.
ANIMAL-10N consists of 50000 train animal images and 10000 test animal images in 10 classes, with an 8\% estimated error rate. 
The temperature is 1, and the coefficient for the regularization term is 2.

For comparison, results of F-correction \cite{patrini2017making}, Decoupling \cite{malach2017decoupling}, D2L \cite{ma2018dimensionality}, MentorNet \cite{jiang2018mentornet}, Co-teaching \cite{han2018co}, Iterative-CV \cite{chen2019understanding}, DivideMix \cite{li2020dividemix}, SELFIE \cite{DBLP:conf/icml/SongK019}, PLC \cite{DBLP:conf/iclr/ZhangZW0021}, NCT \cite{DBLP:conf/cvpr/ChenSHS21} are reported.

As shown in Table~\ref{table_webvision}, Robust LR improves the performance by a considerable margin, namely, 4.5\% top-1 accuracy against the previous best on the test set of Mini-WebVision and 4.4\% on ANIMAL-10N.
The results verify that our method can cope with complex real-world noise.

\subsection{Ablation study}

We further study the components of Robust LR. 
Specifically, we analyze the results of:
\begin{enumerate}[1.]
    \item To study the effect of label refurbishment, we remove label refurbishment and directly use either given noisy labels or pseudo-labels.
        When the probability of being clean is larger than 0.5, the noisy label is used. 
        Otherwise, the pseudo-label is used.
    \item To study the effect of strong augmentation, we replace it with basic transformation.
    \item To study the effect of GMM for dynamic confidence estimation, we replace it with 0.5 fixed confidence.
    \item To study the effect of co-training, we only use one model.
\end{enumerate}

The results on CIFAR-10 with four levels of symmetry noise are reported.
From Table~\ref{table_ab}, other training schemes suffer from different degrees of performance drops.
This verifies that the incorporation of the components in Robust LR is effective.
In the next, we analyze each component.

\subsubsection{Label refurbishment}
The label refurbishment alleviates the marginalized effect of wrong labels and thus, contributes to the final performance.
Under light noise, i.e., when the noise is insignificant or can be corrected easily, the gain is limited.
Under heavy noise, e.g., 90\% noise rate, the model is much more sensitive to its absence.

We also notice the large gap between the best and last performance (73.8\% vs. 23.9\%) under heavy noise. 
To understand, we further observe models' behaviors.
We find that the training is unstable under heavy noise, e.g., the GMM may not converge in some rounds and assigns more than 95\% of examples with bigger clean probabilities.
The bad confidence estimation would affect later training in return.
The training can be stabilized after further tuning the hyper-parameters, such as the learning rate.
For consistency, we only report the performance under the same hyper-parameters.

\begin{table}
    \centering
    \begin{tabular}{ll|cccc}
        \toprule
		\multicolumn{2}{l|}{Method/Noise ratio}                           & 20\% & 50\% & 80\% & 90\%   \\ \midrule
		\multirowcell{2}[0ex][l]{Robust LR}                               & Best & 96.5 & 95.8 & 94.5 & 92.8   \\
		                                                                  & Last & 96.4 & 95.7 & 94.2 & 92.8   \\ \midrule
		\multirowcell{2}[0ex][l]{1. w/o LR}          & Best & 96.3 & 95.8 & 94.5 & 73.8   \\ %
		                                                                  & Last & 96.2 & 95.6 & 94.1 & 23.9   \\ \midrule
		\multirowcell{2}[0ex][l]{2. w/o strong aug.}          & Best & 92.6 & 88.1 & 65.3 & 48.7   \\ %
		                                                                  & Last & 92.5 & 72.7 & 36.5 & 24.3   \\ \midrule
		\multirowcell{2}[0ex][l]{3. w/o GMM}                          & Best & 94.6 & 91.4 & 88.0 & 87.6   \\ %
		                                                                  & Last & 92.7 & 80.3 & 43.4 & 31.1   \\ \midrule
		\multirowcell{2}[0ex][l]{4. w/o co-training}                  & Best & 96.4 & 95.7 & 94.3 & 82.8   \\ %
		                                                                  & Last & 95.6 & 94.4 & 93.1 & 79.9   \\ 
        \bottomrule
    \end{tabular}
    \caption{Ablation study. Results on CIFAR-10 with different levels of symmetry noise are reported.}
    \label{table_ab}
\end{table}

\subsubsection{Data augmentation}
Replacing the strong augmentation is detrimental to performance.
Without it, the model fails to converge.
We remark that it's because Robust LR is a holistic method. 
Strong data augmentation not only serves the common purpose of regularization \cite{shorten2019survey}, but also is part of the different augmentation strategies \cite{nishi2021augmentation}.

One may still argue that the augmentation is more important than other components.
We show that other components all improve upon the Robust LR with strong augmentation in Table \ref{table_ab}.
Besides, as shown in Table \ref{table_CIFAR}, our method outperforms AugDesc, a method with the same augmentation Robust LR uses.

\subsubsection{GMM}
The GMM is also essential, and removing the dynamic confidence estimation damages the performance.
We also notice that, for four levels of corruption, GMM assigns 18\%, 44\%, 70\%, 78\% examples bigger noisy probability ($w<0.5$) at the end of training, respectively.
It is an accurate estimation of the real noise rate (for 20\%, 50\%, 80\%, 90\% noise rate, there is actually 18\%, 45\%, 72\%, 81\% noisy labels).
For Mini-WebVision, the GMM assigns 18\% examples bigger noisy probability in the end, which is also approximate to the reported noise rate 20\% \cite{li2017webvision}.
We envision this could be used to estimate the noise rate in real-world datasets.

\subsubsection{Co-training}
Removing co-training leads to considerable drops in performance.
It is also noteworthy that our single model's performance already surpasses previous co-training methods, such as DivideMix or Co-teaching.
We suggest that co-training alleviates confirmation bias, and the ensemble of two models also produces better self-training signals.

\subsection{Finding the noisy labels in CIFAR-10}
Apart from combating label noise, Robust LR can be directly used to find the noisy labels in the training set.
Standard empirical risk minimization would easily fit the training set with only a small amount of noisy labels.
Instead, Robust LR could avoid the fitting on the possible noisy labels.
We use CIFAR-10 to illustrate how we can use Robust LR to find noisy labels in a mostly correctly labeled dataset.

\begin{figure}
    \centering
    \includegraphics[width=1.0\columnwidth]{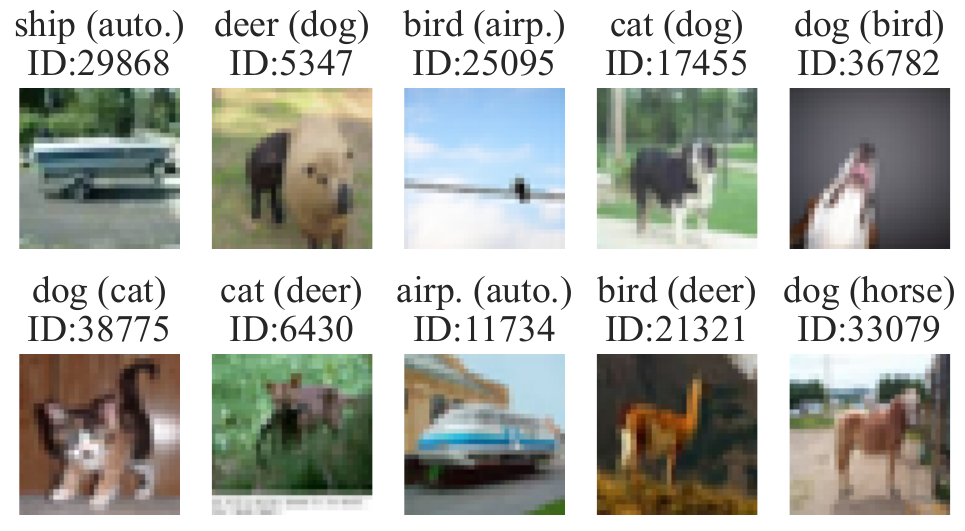} 
    \caption{Some mislabeled or indistinguishable examples in the training set of CIFAR-10 found by Robust LR. 
            The wrong annotations, the predicted classes (in the parentheses), and the IDs of images are shown.
            The airp. and auto. are airplane and automobile for short.
            } 
    \label{labelerror}
\end{figure}

We first train Robust LR on the CIFAR-10 training set (without corruption) for 100 epochs without modifying the algorithm.
In the next, examples with top-50 big losses are selected and hand-picked.
We successfully find some mislabeled or indistinguishable examples in the training set as in Figure~\ref{labelerror} (note that there is no ground-truth or high-resolution originals, we can only subjectively tell whether the noisy labels are right or wrong).
Some of them are mislabeled probably because of the similarity between two classes, such as 25095 (bird vs. airplane), 38775 (dog vs. cat).
Some classes don't usually consider similar, but images in these classes can still be ambiguous, e.g., image 36782 (dog vs. bird) and 33079 (dog vs. horse).
These verify that the noise we are facing in the real world could be complex.

\section{Conclusion}
In this paper, we study the problem of learning with label noise. 
We analyze the drawbacks of the two-stage pipeline and identify its confirmation bias problem by visualizing the training dynamics.
The observation motivates us to propose Robust LR, a new training algorithm that dynamically refurbishes labels using confidence estimation and pseudo-labeling techniques.
We demonstrate that our approach combats both confirmation bias and label noise.
As a result, it significantly advances the state-of-the-art.
We then conduct ablation experiments to study the effects of the components.
Finally, we attempt to find the mislabeled examples in CIFAR-10 with Robust LR. 
In future work, we are interested in further incorporating ideas from weakly supervised learning into hybrid methods and continuing to combat complex label noise. 
\clearpage
\section{Acknowledgements}
This paper is supported by the National Natural Science Foundation of China (Grant No. 62192783, U1811462), the Collaborative Innovation Center of Novel Software Technology and Industrialization at Nanjing University.
\bibliography{egbib}
\appendix
\clearpage
\appendix

\section{Training details}
We implement our model in PyTorch 1.6 (\url{https://github.com/pytorch/pytorch}).
The GMM is fitted using the scikit-learn package (\url{https://scikit-learn.org/}).
We train our model on one NVIDIA V100 GPU.

For CIFAR, the model is trained for 500 rounds after 15 epochs of warm-up.
In every round, we train the network using SGD with a learning rate of 0.03, a momentum of 0.9, a weight decay of 0.0005, a batch size of 448, and iterations of 222 (two loops over the training set).
The learning rate is reduced by a factor of 10 in the last 100 rounds.
The GMM is fitted with a maximal iteration of 10, a convergence threshold of 0.01, a non-negative regularization of 0.0005.
When the noise rate is 90\%, the losses in the last 5 epochs are averaged to stabilize the fitting of GMM.
For GMM, the convergence threshold is 0.01, and the non-negative regularization is 0.0005. Other hyper-parameters follow the default settings of scikit-learn.

For Mini-WebVision, the model is trained for 300 rounds after 1 epoch of warm-up.
In every round, we train the network using SGD with a learning rate of 0.01, a momentum of 0.9, a weight decay of 0.0005, a batch size of 160, and iterations of 1000.
The learning rate is reduced by a factor of 10 in the last 100 rounds.
For GMM, the convergence threshold is 0.01, the non-negative regularization is 0.001, and other hyper-parameters follow the default settings of scikit-learn.
The model is the inception-resnet v2 \cite{szegedy2017inception}.
For ANIMAL-10N, the model is trained for 500 rounds after 15 epochs of warm-up.
In every round, we train the network using SGD with a learning rate of 0.01, a momentum of 0.9, a weight decay of 0.0005, a batch size of 64, and iterations of 1564 (two loops over the training set).
The learning rate is reduced by a factor of 10 in the last 100 rounds.
The model is the VGG-19 \cite{DBLP:journals/corr/SimonyanZ14a}.
For GMM, the convergence threshold is 0.01, and the non-negative regularization is 0.0005. Other hyper-parameters follow the default settings of scikit-learn.

\section{Training curve}
\begin{figure}[h!]
    \centering
    \includegraphics[width=\columnwidth]{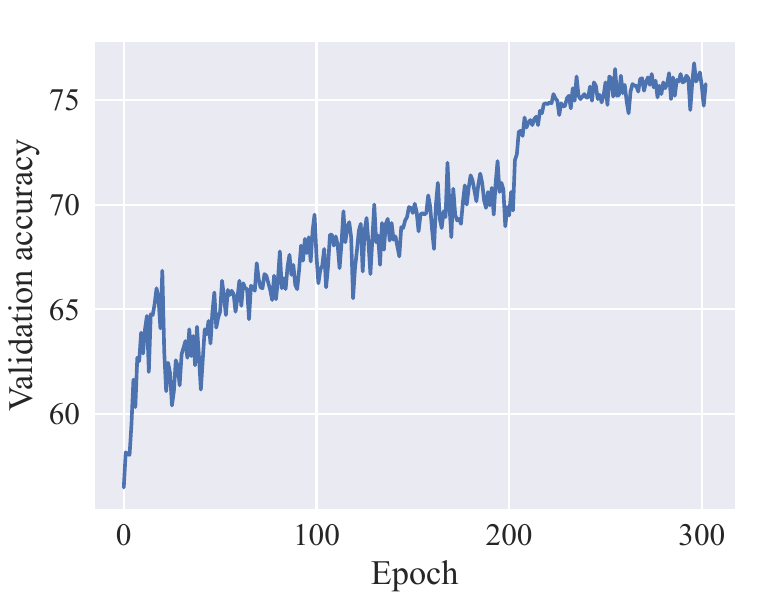} 
    \caption{Training curve on the Mini-WebVision's validation set. The results on noisy training set is not shown because the accuracy on noisy dataset couldn't reflect model's real performance.} 
    \label{fig_curve}
\end{figure}

The training curve on Mini-WebVision is shown in Fig. \ref{fig_curve}.

\section{Details of transformations}
\label{Details of transformations}
The strong transformation is a modified version of RandAugment \cite{cubuk2020randaugment} followed by Cutout \cite{devries2017improved}.
It basically follows the setting of FixMatch \cite{sohn2020fixmatch}.
The operations of RandAugment are shown in Table~\ref{strongaug}.
The meaning of range is the same as the original version, so we don't elaborate here.
Cutout randomly masks a square (with a side of length ranging from 0 to 0.5×image length) of pixels to gray. 

\begin{table}[H]
    \centering
    \caption{
        List of operations for strong transformations of the modified RandAugment. 
        Three transformations are randomly chosen and performed with stochastic magnitude.
    }
    \label{strongaug}
    \begin{tabular}{ll|ll}
        \toprule
        Operation      & Range        & Operation      & Range        \\ \midrule
        AutoContrast   & [0, 1]       & Rotate         & [-30, 30]    \\
        Brightness     & [0.05, 0.95] & Sharpness      & [0.05, 0.95] \\
        Color          & [0.05, 0.95] & ShearX         & [-0.3, 0.3]  \\
        Contrast       & [0.05, 0.95] & ShearY         & [-0.3, 0.3]  \\
        Equalize       & [0, 1]       & Solarize       & [0, 256]     \\
        Identity       & [0, 1]       & TranslateX     & [-0.3, 0.3]  \\
        Posterize      & [4, 8]       & TranslateY     & [-0.3, 0.3]  \\
        \bottomrule
    \end{tabular}
\end{table}

\end{document}